\documentclass[10pt,twocolumn,letterpaper]{article}

\usepackage[pagenumbers]{iccv} % To force page numbers, e.g. for an arXiv version

\usepackage{colortbl}
\usepackage{marvosym} % 在导言区添加

\definecolor{iccvblue}{rgb}{0.21,0.49,0.74}
\usepackage[pagebackref,breaklinks,colorlinks,allcolors=iccvblue]{hyperref}
\usepackage{multirow}

\newcommand{\ModelName}{ImageGen-CoT\xspace}

\def\paperID{13708} % *** Enter the Paper ID here
\def\confName{ICCV}
\def\confYear{2025}

\title{ImageGen-CoT: Enhancing Text-to-Image In-context Learning with Chain-of-Thought Reasoning}

\author{Jiaqi Liao$^{1\dag}$, Zhengyuan Yang$^{1}$, Linjie Li$^{1}$, Dianqi Li, Kevin Lin$^{1}$, Yu Cheng$^{2}$, Lijuan Wang$^{1\textsuperscript{\Letter}}$\\
{\normalsize\centering$^{1}$ Microsoft}
{\normalsize\centering$^{2}$ The Chinese University of Hong Kong}
}

\begin{document}
\twocolumn[{%
    \renewcommand\twocolumn[1][]{#1}%
    \maketitle
    \centering
    \captionsetup{type=figure}
    \includegraphics[width=1.0\linewidth]{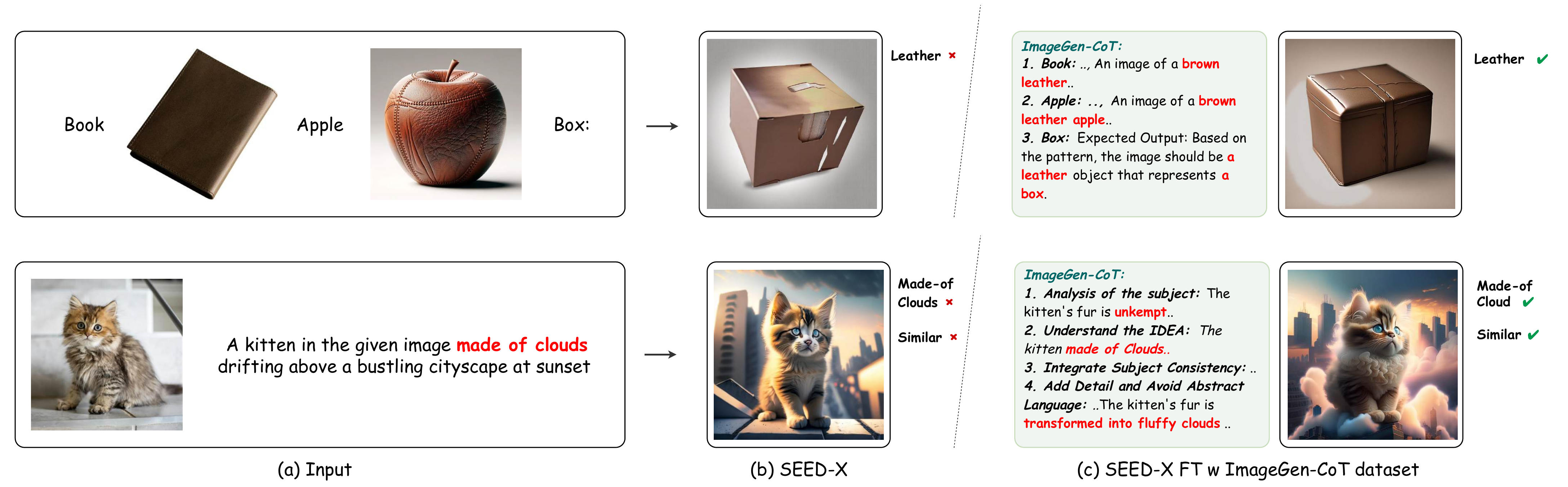}
    \caption{\textbf{Comparisons between SEED-X and SEED-X FT w ImageGen-CoT dataset.} Our method forces the model to generate a thought process before the ImageGen. The top row shows SEED-X failing to infer ‘Leather’ style, generating only a box, while our approach enables SEED-X to recognize the leather style and produce the intended leather box. The bottom row shows SEED-X failing to capture the unkempt fur and cloud, while our method successfully recognizes these key elements.}
    \vspace{0.3cm}
    \label{fig:teaser}
}]

{\let\thefootnote\relax\footnote{\dag~Interns at Microsoft.}}
\begin{abstract}
\vspace{+3pt}
% \zyang{TODO: list 1-2 sentences for each story's core contribution. Then enumerate all experiments (short bullets) that perfectly support the main claim, and concerns/flaws to complete that story.}

In this work, we study the problem of Text-to-Image In-Context Learning (T2I-ICL). While Unified Multimodal LLMs (MLLMs) have advanced rapidly in recent years, they struggle with contextual reasoning in T2I-ICL scenarios. To address this limitation, we propose a novel framework that incorporates a thought process called ImageGen-CoT prior to image generation. 
% However, we observe that MLLMs often produce unstructured reasoning steps, resulting in suboptimal outcomes. To tackle this issue,
To avoid generating unstructured ineffective reasoning steps, we develop an automatic pipeline to curate a high-quality ImageGen-CoT dataset. We then fine-tune MLLMs using this dataset to enhance their contextual reasoning capabilities. 
% However, due to the complexity of T2I-ICL tasks, there is still significant room for improvement. 
To further enhance performance, we explore test-time scale-up strategies and propose a novel hybrid scaling approach. This approach first generates multiple ImageGen-CoT chains and then produces multiple images for each chain via sampling. Extensive experiments demonstrate the effectiveness of our proposed method. Notably, fine-tuning with the ImageGen-CoT dataset leads to a substantial 80\% performance gain for SEED-X on T2I-ICL tasks. See our project page at \url{https://ImageGen-CoT.github.io/}. Code and model weights will be open-sourced.

\end{abstract}      
\section{Introduction}
\label{sec:intro}
Human intelligence excels at learning novel concepts through contextual observation and adapting to new inputs. When presented with a series of interleaved text-image examples—such as ``a leather-bound book'', followed by ``a leather apple''—and then asked to generate an image for the query ``a box,'' humans intuitively infer the implicit pattern of ``leather'' and apply it to the new query, resulting in ``a leather box''. This reasoning ability to learn novel concepts from multimodal contexts underpins creative problem-solving.  Existing unified Multimodal Large Language Models (unified MLLMs)~\cite{ge2023making,dong2023dreamllm, sun2024generative, lu2024unified, ge2024seed, wang2024emu3,li2024multimodal} have demonstrated remarkable capabilities in multimodal understanding and generation within a single model architecture. Given their ability to process and generate across modalities similar to human cognition, it is natural to investigate whether these models can exhibit reasoning capabilities comparable to those of humans. To evaluate this, we adopt the Text-to-Image In-Context Learning (T2I-ICL) task~\cite{zeng2024can}, which requires models to process interleaved text-image inputs and generate coherent outputs by learning from multimodal contexts (Figure~\ref{fig:teaser}). Despite the impressive capabilities of unified MLLMs, our experiments reveal that they struggle to replicate this reasoning capability, often failing to grasp contextual relationships or preserve compositional consistency in T2I-ICL tasks.

To overcome these challenges, building upon the demonstrated success of CoT prompting in enhancing complex task processing for LLMs, we propose a novel framework that involves a structured thought process called \textbf{ImageGen-CoT} prior to image generation. Our key insight is that explicitly \textbf{generating reasoning steps before image synthesis} helps unified MLLMs better understand multimodal contexts and produce more coherent outputs. However, these models often produce disorganized and incoherent thought processes, leading to suboptimal performance. To address these limitations, we first propose an automated dataset construction pipeline to generate ImageGen-CoT datasets, where each sample consists of a pair of ImageGen-CoT and a corresponding image. The pipeline comprises three main stages: 1) collecting T2I-ICL instructions, 2) using MLLMs to generate step-by-step reasoning (ImageGen-CoT), and 3) producing image descriptions via MLLMs for diffusion models to generate images. To further enhance the dataset quality, we employ an iterative refinement process: The model first generates multiple text prompts and corresponding images, selects the best one, critiques the generated image, and iteratively refines the prompt until the max round is reached.
\textbf{Then, we fine-tune the model using this dataset} which significantly enhances the image generation capabilities of unified-MLLMs in T2I-ICL tasks. 

Despite the strong performance, T2I-ICL tasks' complexity leaves room for improvement. Inspired by NLP's Best-of-N paradigm, we explore \textbf{three test-time scaling strategies}:
1. Multi-Chain: Generate multiple ImageGen-CoT chains, each producing one image; 
2. Single-Chain: Create multiple image variants from one ImageGen-CoT; 
3. \textbf{Hybrid:} Combine both methods - multiple reasoning chains with multiple image variants per chain. Our empirical studies reveal two critical insights: (1) 	Instead of changing seeds, generating multiple ImageGen-CoTs via high-temperature LLM decoding achieves a similar performance to scaling. (2) \textbf{ImageGen-CoT enables bidirectional expansion}—either generating multiple instances of ImageGen-CoT or modifying seeds to create diverse images—outperforming single-dimension scaling, opening new pathways for performance optimization in complex multimodal tasks.

\begin{figure}[t]
    \centering
    \includegraphics[width=1.02\columnwidth]{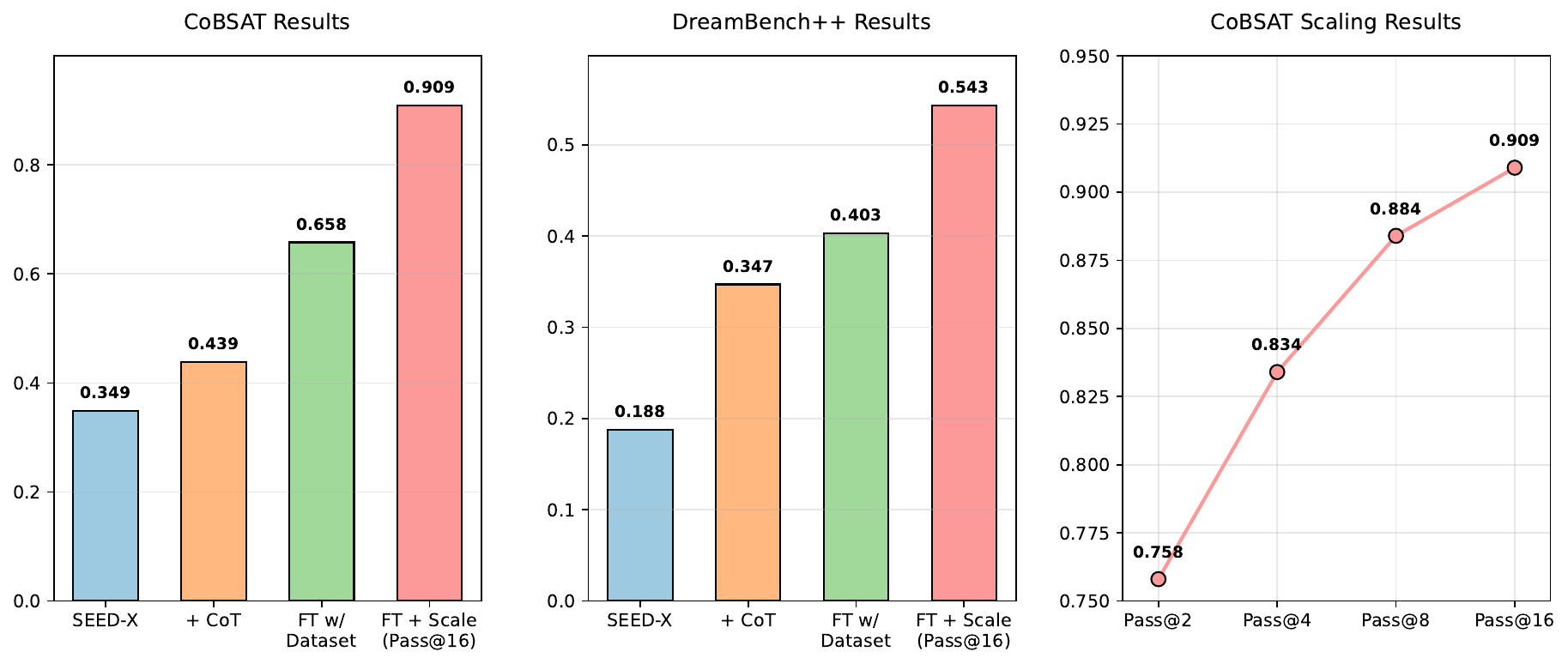}
    \vspace{-1.0mm}
    \caption{
        \textbf{Performance comparison on CoBSAT and DreamBench++ benchmarks.} Our method significantly improves SEED-X's performance through progressive enhancements: adding ImageGen-CoT, fine-tuning with the ImageGen-CoT dataset, and applying test-time scaling strategies. 
    }
    \vspace{-1.5mm}
    \label{fig:teaser_result}
\end{figure}

To evaluate the effectiveness of our method, we experiment with leading Unified MLLMs. These models can be categorized into two types based on their visual representations: discrete visual tokens~\cite{ge2023making, lu2024unified, team2024chameleon, wang2024emu3, qu2024tokenflow} and continuous visual embeddings~\cite{dong2023dreamllm, tong2024metamorph, xie2024show, li2024synergen}. We select SEED-LLaMA~\cite{ge2023making} as a representative of the discrete approach and SEED-X~\cite{ge2024seed} for the continuous approach, considering their open-source availability and support for interleaved text-image input. Extensive experiments demonstrate the
effectiveness of our method. Specifically, as shown in Figure~\ref{fig:teaser_result}, SEED-X FT with ImageGen-CoT improves by 89\% and 114\% on CoBSAT and DreamBench++. With scaling strategy, it further achieves 0.909 and 0.543 respectively.

Our contributions can be summarized as follows:
\begin{enumerate}
\item We propose a novel framework that generates a thought process (called ImageGen-CoT) to enhance the performance of unified MLLMs in T2I-ICL tasks. \\
\item We construct high-quality \ModelName datasets for fine-tuning unified MLLMs through an automatic dataset construction pipeline. \\
% \item To further improve the performance, 
\item We explore Best-of-N test-time scaling up paradigms and propose a hybrid scaling approach that first generates multiple ImageGen-CoT chains and then generates multiple image variations per chain.

\end{enumerate}

\section{Related Work}
\label{sec:related}
\subsection{In-Context Learning}
% As first proposed by Brown et al. (2020)~\cite{brown2020language}, l
Large language models (LLMs)~\cite{brown2020language} have exhibited exceptional capabilities for text in-context learning (T-ICL). 
This ability allows LLMs to adapt to new tasks by observing a few illustrative examples provided as context, without requiring parameter updates. 
These models demonstrate remarkable performance on various tasks, with extension to multimodal models~\cite{tsimpoukelli2021multimodal,alayrac2022flamingo,yang2022empirical}.
With the development of image generation, recent studies have proposed text-to-image in-context learning (T2I-ICL). 
 For instance, CoBSAT~\cite{zeng2024can} is the first benchmark designed to evaluate a model’s T2I-ICL (Text-to-Image In-Context Learning) generation capabilities. This includes assessing the model’s ability to rapidly adapt to tasks given the in-context demonstrations, which are key aspects of T2I-ICL.
 Emu2~\cite{sun2024generative} also evaluates models’ T2I-ICL capabilities through subject customization in DreamBench~\cite{ruiz2023dreambooth}, where the model needs to bind visual concepts from reference images to generate customized outputs. In this work, following previous studies, we validate our approach’s improvement on the T2I-ICL task using CoBSAT and DreamBench++ \cite{peng2024dreambench++}. % (an extension of DreamBench).

\subsection{Text-to-Image Generation}
Text-to-Image (T2I) generation~\cite{rombach2022high,ramesh2022hierarchical,yu2022scaling,saharia2022photorealistic,podell2023sdxl} aims to generate images based on a user’s textual description. With the development of T2I diffusion models, such as DALL-E 3~\cite{betker2023improving}, SD3~\cite{esser2024scaling}, and FLUX.1-Schnell~\cite{flux2023}, users can now generate high-quality and vivid images directly from text descriptions. Building on this success, there is an increasing demand for models to generate customized content, such as specific subjects, styles, or attributes tailored to individual user needs. Consequently, a variety of methods have emerged to address the challenge of subject-customized generation. These methods~\cite{gal2022image,kumari2023multi, ruiz2023dreambooth, park2024cat, sohn2024styledrop} typically rely on fine-tuning techniques, such as LoRA~\cite{hu2021lora} or contrastive learning~\cite{he2020momentum}, to specialize a general T2I model for subject customization. However, these methods require the collection of subject-specific datasets and involve time-consuming retraining for each new user request. This makes them resource-intensive, limiting their ability to generalize quickly to new needs. To address these challenges, researchers~\cite{sun2024generative} train EMU2 on multimodal sequences, leveraging its inherent ICL ability to quickly bind visual concepts from the context. Despite these advancements, their performances remain limited. In this work, we explore how introducing a thought process prior to image generation, called ImageGen-CoT, can significantly enhance their performance on the T2I-ICL task.

\subsection{Unified Multimodal Language Models}
Recent years have witnessed remarkable progress in multimodal AI across two key domains: understanding and generation. In understanding, Large Vision-Language Models (LVLMs)~\cite{alayrac2022flamingo, liu2024improved, zhu2023minigpt, lu2024deepseek, chen2024far, yang2024qwen2,wang2022git,yang2023dawn,li2024multimodal} have achieved impressive capabilities in complex visual-textual reasoning tasks. Meanwhile, in generation, Text-to-Image diffusion models~\cite{betker2023improving, esser2024scaling, flux2023} have advanced to produce photorealistic images that can rival professional artists' work. Given these developments, researchers have been exploring ways to unify multimodal understanding and generation capabilities within a single model architecture. These models can be categorized into two approaches based on their visual representations: discrete visual tokens~\cite{ge2023making, lu2024unified, team2024chameleon, wang2024emu3, qu2024tokenflow} and continuous visual embeddings~\cite{dong2023dreamllm, tong2024metamorph, xie2024show, li2024synergen}. Discrete approaches leverage VQ-VAE to tokenize images into discrete tokens, enabling training and inference similar to language processing. In contrast, continuous approaches generate latent embeddings that are subsequently processed through diffusion models for image synthesis.
\section{Method}
\label{sec:method}

\begin{figure*}[h]
    \centering
    \includegraphics[width=\textwidth]{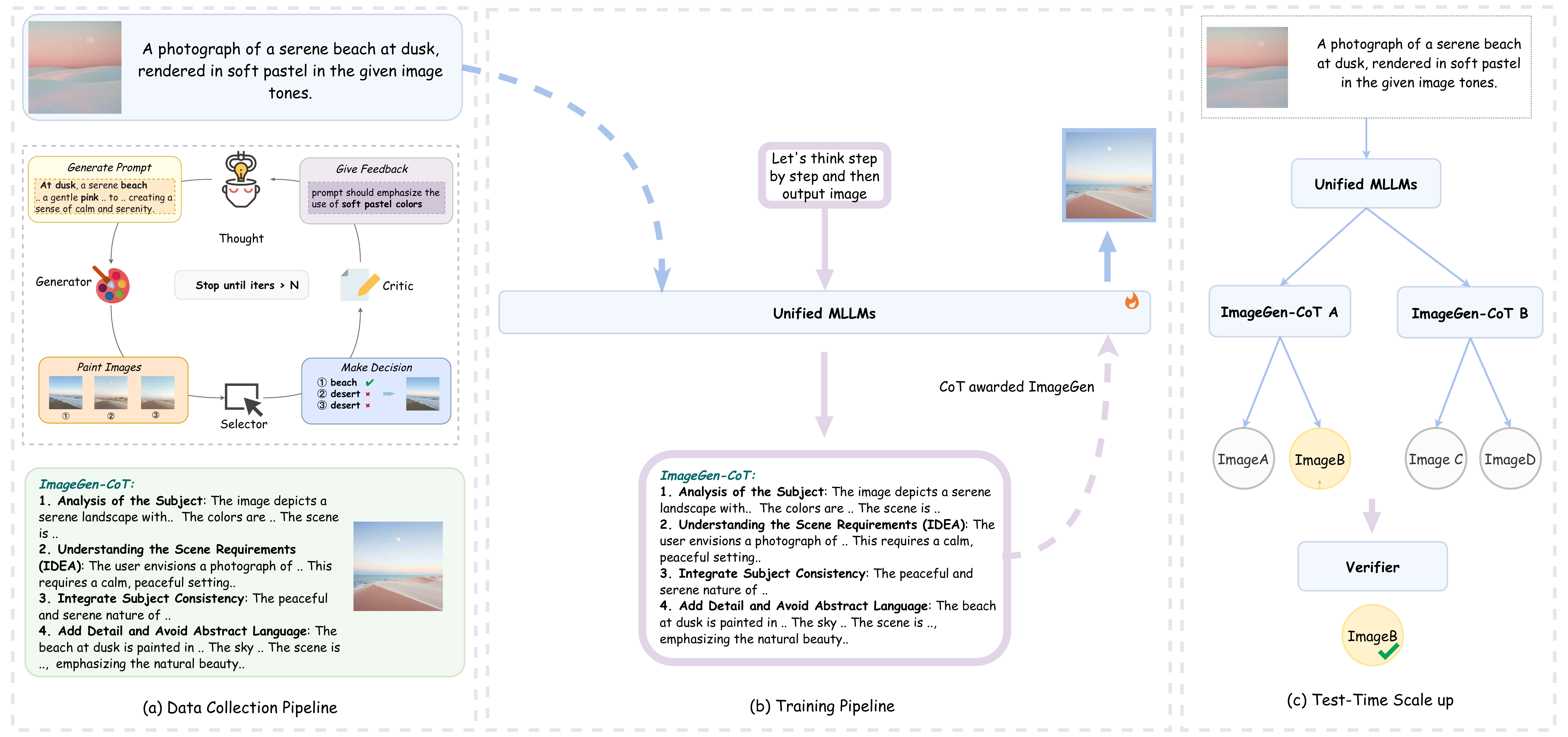}
    \vspace{-0.3mm}
    \caption{
        \textbf{Main Pipeline.} (a) \textbf{Data Collection Pipeline}: An automated iterative process where the MLLM acts as Generator, Selector, Critic, and Refiner to produce high-quality ImageGen-CoT (reasoning chains) and aligned images. (b) \textbf{Training Pipeline}: Fine-tuning unified MLLMs on the collected ImageGen-CoT dataset to enhance contextual reasoning and image generation. (c) \textbf{Test-Time Scaling}: Strategies for performance improvement via hybrid scaling during inference.
    }
    \vspace{-0.3mm}
    \label{fig:main_pipeline}
\end{figure*}

In this section, we present our \ModelName framework in detail. First, we introduce the formulation of \ModelName. (Sec.\ref{sec:method}). Second, we describe our automated pipeline for collecting high-quality ImageGen-CoT datasets (Sec.\ref{sec:dataset}). Third, we provide a detailed formulation of the dataset and the loss function used to fine-tune the model with the collected dataset (Sec.\ref{sec:training}). Finally, we explore various strategies to enhance model performance during inference, culminating in a novel hybrid scaling approach that addresses both contextual comprehension and generation challenges (Sec.\ref{sec:scale}).

\subsection{Formulation of \ModelName}
~\label{sec:method}
As described above, T2I-ICL tasks require models to have a high level of comprehension. To enhance the model’s capacity, we propose a new framework that generates a Chain-of-Thought, which we call ImageGen-CoT, before performing ImageGen. While we initially expected models to simultaneously output both ImageGen-CoT reasoning chains and corresponding images in a single forward pass. 

However, during our practice, we observe that models frequently fail to generate images even when explicitly prompted to first generate ImageGen-CoT followed by image output. As illustrated in Figure~\ref{fig:main_pipeline}, to ensure reliable image generation, we develop a two-stage inference protocol. The first stage involves prompting the model to generate the ImageGen-CoT reasoning chain $R$. In the second stage, we combine the original input $X$ with the generated ImageGen-CoT $R$, along with a mandatory image generation token $\langle \text{image} \rangle$, to guarantee the production of the target image $I$. This process can be formally expressed as:

\begin{equation}
\begin{aligned}
& \text{Stage 1: } R = \mathcal{M}(X \oplus \text{instruction}) \\
& \text{Stage 2: } I = \mathcal{M}(X \oplus R \oplus \langle \text{image} \rangle)
\end{aligned}
\end{equation}

where $\mathcal{M}$ denotes the unified MLLMs, and $\oplus$ represents the concatenation operation. 

\subsection{Dataset Construction}
~\label{sec:dataset}
Due to the limitations of some unified MLLMs in generating well-structured ImageGen-CoT, which leads to suboptimal performance, we propose an automated pipeline to collect an ImageGen-CoT dataset and fine-tune the model using this dataset.

To collect high-quality \ModelName datasets, we first establish an instruction pool by collecting instructions from existing training datasets in T2I-ICL tasks. Second, we propose an automatic dataset construction pipeline. As illustrated in Figure~\ref{fig:main_pipeline}, our pipeline proceeds as follows: In the initial stage, we let MLLM act as a \textbf{Generator} to generate N outputs, each consisting of an ImageGen-CoT and a prompt for the next image, which are then used by T2I-Model~\cite{flux2023} to generate N images. Then, MLLM acts as a \textbf{Selector} to select the best image from the N candidates. After that, if the selected image meets our quality threshold or reaches the maximum iteration limit, the pipeline terminates and outputs the corresponding ImageGen-CoT and image pair. Otherwise, we let MLLM act as a \textbf{Critic} to write a critique of the selected image, assessing how well it matches the T2I-ICL prompt. Finally, MLLM acts as a \textbf{Refiner} to refine the prompt based on the critique, and the process iterates until meeting the termination. Finally, based on the collected responses, we construct our ImageGen-CoT dataset as follows:

\begin{equation}
\mathcal{D}_{\text{ImageGen-CoT}} = \{(T_i^*, I_i^*)\}_{i=1}^n
\end{equation}
where $(T_i^*, I_i^*)$ represents the $i$-th high-quality pair selected by our pipeline. $T_i^*$ is the ImageGen-CoT that successfully guided the generation. $I_t^*$ is the corresponding generated image that meets our quality standards. $n$ is the total number of collected pairs in the dataset.

\subsection{ Training Pipeline}
~\label{sec:training}

After constructing the dataset, we explore the training paradigm to fine-tune Unified MLLMs using our collected dataset. In this section, we detail the process of training Unified MLLMs with the ImageGen-CoT dataset, focusing on data formulation and training objectives.

To maintain consistency between the training and inference stages, we divide the ImageGen-CoT dataset into two splits:

(1) $[X, p_\text{cot}] \to \text{ImageGen-CoT}$, which generates the ImageGen-CoT

(2) $[X, \text{ImageGen-CoT}, p_\text{image}] \to \text{image}$, which generates the final image.

When training with the training dataset split 1, since model only generate the ImageGen-CoT text, we apply the normal $lm\_loss$, formulated as follows:

\[
lm\_loss = - \frac{1}{N} \sum_{i=1}^{N} \log P(y_i \mid y_{<i}, X)
\]

where $y_i$ is the $i$-th token in the ImageGen-CoT text, $y_{<i}$ represents the preceding tokens, $X$ is the input, and $N$ is the total number of tokens in the ImageGen-CoT sequence.

For training with dataset split 2, there is no uniform training loss, as different Unified MLLMs utilize varying visual representations (e.g., discrete visual tokens~\cite{ge2023making, wang2024emu3} or continuous visual embeddings~\cite{ge2024seed}). For models using discrete visual tokens, the same loss as language modeling ($lm\_loss$) is applied. For models using continuous visual embeddings, we apply the $mse\_loss$ between the generated and target visual embeddings, formulated as:
\[
mse\_loss = \|\hat{z} - z\|^2
\]

where $\hat{z}$ is the generated visual embedding and $z$ is the corresponding target visual embedding. In this study, our primary objective is to enhance the model's capability to generate accurate ImageGen-CoT. So, by default, we utilize data split 1 for fine-tuning Unified MLLMs, with more results presented in the Appendix.

\subsection{Test time scale up}
\label{sec:scale}

Though fine-tuning with the ImageGen-CoT dataset significantly improves model performance in T2I-ICL tasks, substantial room for improvement remains. Inspired by test-time scaling methods in NLP, we explore whether increasing computational investment during inference can further enhance performance.  We first investigate a conventional paradigm: using SEED-X as the base model, generating multiple images by varying the seed value, and outputs are filtered via a ground-truth verifier aligned with the Pass@N metric. However, we observe that even with N=16, this approach underperforms compared to SEED-X fine-tuned with ImageGen-CoT Dataset. 

This observation motivates our exploration of test-time scaling in the context of ImageGen-CoT, which we approach through three distinct strategies: 1. \textbf{Single-Chain Scaling:} This approach generates one ImageGen-CoT chain and produces multiple image variants by varying the seed values. 2. \textbf{Multi-Chain Scaling:} Similar to NLP's "Best-of-N" sampling, we generate multiple ImageGen-CoT chains through high-temperature LLM decoding. Each chain produces a unique image, potentially capturing different aspects of the contextual requirements. 3.\textbf{Hybrid Scaling:} Regarding the dual challenges of contextual comprehension and generation in T2I-ICL tasks, we propose a hybrid approach that combines the strengths of both strategies. As illustrated in Figure~\ref{fig:main_pipeline}, this method first generates multiple ImageGen-CoT chains and then creates multiple image variations for each chain. Our extensive experiments further reveal the effectiveness of this hybrid scaling strategy: the integration of ImageGen-CoT enables effective bidirectional scaling across both comprehension and generation dimensions. This dual-axis scalability opens new pathways for optimizing MLLM performance in complex multimodal tasks.

\section{Experiments}
\label{sec:exp}

% \zyang{TODO: organize main experiments here; as bullets}

\subsection{Implementation details}
To validate the effectiveness of our ImageGen-CoT framework and dataset, we conduct experiments on two T2I-ICL benchmarks: CoBSAT~\cite{zeng2024can} and DreamBench++~\cite{peng2024dreambench++}. We employ SEED-LLaMA~\cite{ge2023making} and SEED-X~\cite{ge2024seed} as our base Unified MLLMs for both ImageGen-CoT reasoning and image generation. For the dataset construction pipeline, we utilize different configurations: on DreamBench++, InternVL2.5-78B-MPO-AWQ~\cite{wang2024mpo} serves as the Generator, Selector, Critic, and Refiner, while for CoBSAT, we implement a self-consistency selector method with other components remaining the same. FLUX.1-schnell~\cite{flux2023} is selected as the base T2I model for both benchmarks. We maintain CoBSAT's original split strategy, while implementing an image-level split for DreamBench++ to ensure no subject overlap. During data construction, we generate 3 outputs per query using a sampling temperature of 0.7 and top-p of 0.8, with a maximum of 2 iterations. Additional details regarding dataset splits, training procedures, and ablation studies are provided in the Appendix.

\subsection{Main Results} 
In this section, we seek to answer the following questions: a) How much the \ModelName improves model's performance (via prompting)? b) To what extent does the performance of the model improve after fine-tuning with the \ModelName dataset? c) Can we invest more time in inference time to improve the performance? Finally, to better demonstrate the effectiveness of our method, we present visible comparison results.

\begin{table*}[t]
    \centering
    \caption{\textbf{Main results on CoBSAT benchmark.} "FT w/ GT Image" denotes fine-tuning with ground truth images, while "FT w/ ImageGen-CoT" represents fine-tuning with our ImageGen-CoT dataset. The results demonstrate that ImageGen-CoT significantly improves model performance, with relative improvements over baseline model shown in red.}
    \vspace{+10pt}
    \resizebox{\linewidth}{!}{
    \begin{tabular}{l|c c c c c|c c c c c|l}
        \toprule
        \multirow{2}{*}{\textbf{Method}} & \multicolumn{5}{c|}{\textbf{Object-Inference Task}} & \multicolumn{5}{c|}{\textbf{Attribute-Inference Task}} & \multirow{2}{*}{\textbf{Avg.↑}} \\
        \cmidrule(lr){2-6} \cmidrule(lr){7-11}
        & Color-I & Bkg-I & Style-I & Action-I & Texture-I & Color-II & Bkg-II & Style-II & Action-II & Texture-II & \\
        \midrule
        SEED-LLaMA & .616 & .216 & .272 & .592 & .112 & .088 & .168 & .192 & .220 & .056 & .254 \\
        + \ModelName (via Prompt) & .700 & .276 & .300 & .408 & .084 & .176 & .292 & .272 & .192 & .132 & .283 \\
        + FT w/ GT Image & .632 & .272 & .352 & .540 & .128 & .164 & .200 & .256 & .172 & .112 & .283 \\
        \rowcolor{gray!15} + FT w/ \ModelName Dataset & .620 & .368 & .384 & .424 & .060 & .192 & .288 & .208 & .216 & .148 & \textbf{.291} {\color{red}↑14.6\%} \\
        \midrule
        SEED-X & .796 & .412 & .316 & .596 & .240 & .176 & .344 & .260 & .252 & .104 & .349 \\
        + \ModelName (via Prompt) & .724 & .440 & .660 & .784 & .216 & .312 & .472 & .228 & .320 & .240 & .439 \\
        + FT w/ GT Image & .936 & .712 & .896 & .860 & .468 & .280 & .324 & .388 & .636 & .424 & .592 \\
        \rowcolor{gray!15} + FT w/ \ModelName Dataset & .884 & .692 & .928 & .936 & .420 & .504 & .612 & .660 & .524 & .424 & \textbf{.658} {\color{red}↑88.5\%} \\
        \bottomrule
    \end{tabular}
    }
    \label{tab:main_result_on_cobsat}
\end{table*}

\begin{table*}[t]
\caption{\textbf{Evaluation results on Dreambench++ benchmark.} CP refers to concept preservation and PF refers to prompt following metrics. "FT" stands for fine-tuning. The relative gains over baseline model are shown in red.}
\vspace{2pt}
\resizebox{\linewidth}{!}{
\begin{tabular}{l|c c c c c|c c c c|l}
\toprule
\multirow{2}{*}{\textbf{Method}} & \multicolumn{5}{c|}{\textbf{Concept Preservation}} & \multicolumn{4}{c|}{\textbf{Prompt Following}} & \multirow{2}{*}{\textbf{CP$\cdot$PF↑}} \\
\cmidrule(lr){2-6}
\cmidrule(lr){7-10}
& \texttt{Animal} & \texttt{Human} & \texttt{Object} & \texttt{Style} & Overall & \texttt{Photorealistic} & \texttt{Style} & \texttt{Imaginative} & Overall & \\
\midrule
SEED-LLaMA & .436 & .315 & .288 & .381 & .358 & .306 & .202 & .154 & .218 & .078 \\
+ \ModelName (via Prompt) & .390 & .241 & .262 & .346 & .317 & .291 & .211 & .170 & .222 & .078 \\
\rowcolor{gray!15} + FT w/ \ModelName Dataset & .399 & .290 & .271 & .318 & .325 & .348 & .355 & .210 & .310 & \textbf{.101} {\color{red}↑29.5\%} \\
\midrule
SEED-X & .647 & .420 & .526 & .571 & .559 & .346 & .342 & .303 & .337 & .188 \\
+ \ModelName (via Prompt) & .547 & .293 & .369 & .424 & .427 & .862 & .775 & .737 & .817 & .347 \\
\rowcolor{gray!15} + FT w/ \ModelName Dataset & .549 & .410 & .403 & .432 & .458 & .922 & .851 & .846 & .881 & \textbf{.403} {\color{red}↑114.4\%} \\
\bottomrule
\end{tabular}
}
\label{tab:main_results_on_dreambench++}
\end{table*}

\begin{figure*}[t]
    \centering
    \includegraphics[width=\textwidth]{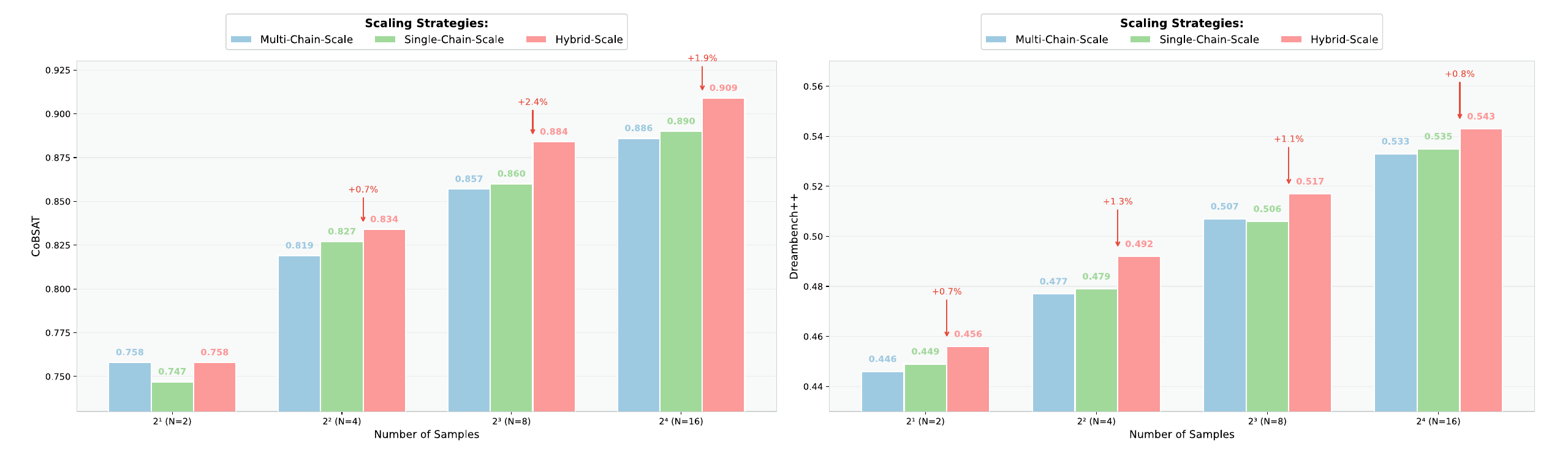}
    \vspace{-5.0mm}
    \caption{\textbf{Test-time scaling strategies comparison.} We conducted a comprehensive evaluation of three distinct scaling strategies: Multi-Chain Scaling, Single-Chain Scaling, and Hybrid Scaling, examining their performance across varying numbers of generated outputs (N=2,4,8,16). The experimental results are presented in two figures, with the left figure showing results on CoBSAT and the right figure displaying results on Dreambench++. The red numbers indicate the performance improvements achieved by Hybrid Scaling compared to Single-Chain Scaling.}
    \vspace{-0.5mm}
    \label{fig:test-time-scale-up}
\end{figure*}

\begin{figure*}[t]
    \centering
    \includegraphics[width=\textwidth]{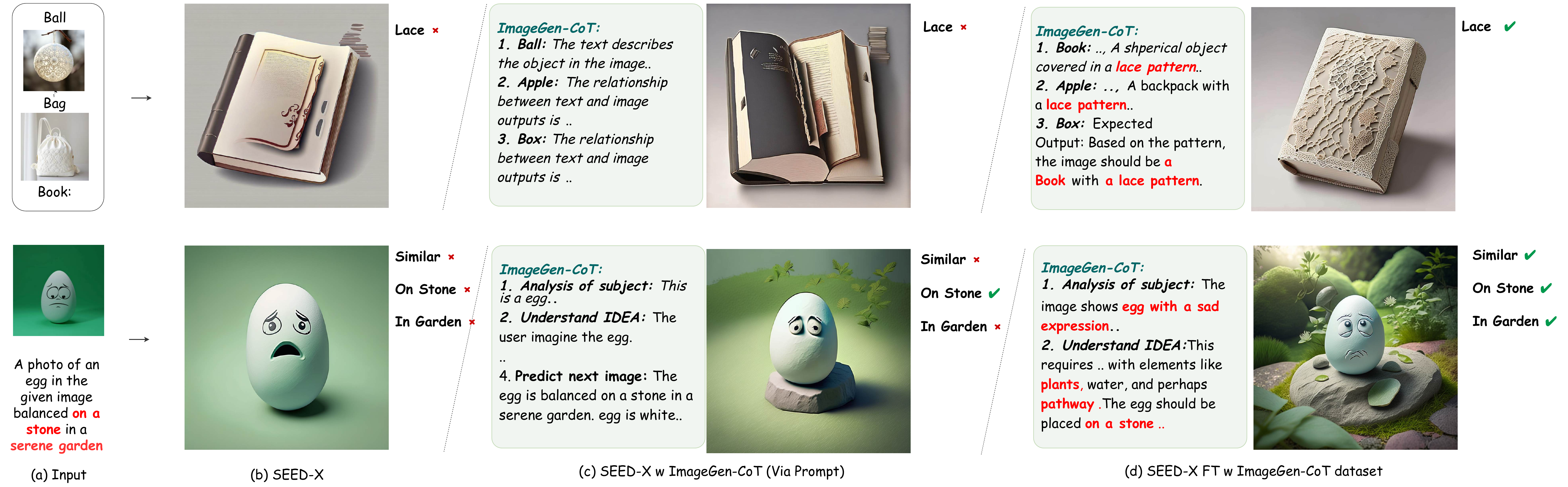}
    \vspace{-3mm}
    \caption{\textbf{Qualitative Results.} Comparison of generation results on COBSAT (top) and Dreambench+ (bottom) using baseline SEED-X, SEED-X with ImageGen-CoT prompting, and SEED-X fine-tuned with ImageGen-CoT dataset.}
    \label{fig:visible_result_on_cobsat_and_dreambench++}
    \vspace{-3mm}
\end{figure*}

\begin{table*}[t]
    \centering
    \caption{We presents a comprehensive analysis of model performance on the COBSAT benchmark. Each model is evaluated in two generation modes: image generation (Img) and text generation (Txt).}
    \resizebox{\linewidth}{!}{
    \begin{tabular}{l|cc|ccccc|ccccc|l}
        \toprule
        \multirow{2}{*}{\textbf{Method}} & \multicolumn{2}{c|}{\textbf{Gen Mode}} & \multicolumn{5}{c|}{\textbf{Object Inference Task}} & \multicolumn{5}{c|}{\textbf{Attribute Inference Task}} & \multirow{2}{*}{\textbf{Avg.↑}} \\
        \cmidrule(lr){2-3} \cmidrule(lr){4-8} \cmidrule(lr){9-13}
        & Img & Txt & Color & Bkg & Style & Action & Texture & Color & Bkg & Style & Action & Texture & \\
        \midrule
        \multirow{2}{*}{SEED-X} 
        & \checkmark & -- & .796 & .412 & .316 & .596 & .240 & .176 & .344 & .260 & .252 & .104 & .349 \\
        & -- & \checkmark & .440 & .388 & .096 & .080 & .060 & .116 & .080 & .180 & .164 & .132 & .174 \\
        \midrule
        \multirow{2}{*}{+ ImageGen-CoT (via Prompt)} 
        & \checkmark & -- & .724 & .440 & .660 & .784 & .216 & .312 & .472 & .228 & .320 & .240 & .439 \\
        & -- & \checkmark & .744 & .212 & .648 & .476 & .388 & .356 & .780 & .292 & .540 & .136 & .457 \\
        \midrule
        \multirow{2}{*}{+ FT w/ ImageGen-CoT Dataset}
        & \cellcolor{gray!15}\checkmark & \cellcolor{gray!15}-- & \cellcolor{gray!15}.884 & \cellcolor{gray!15}.692 & \cellcolor{gray!15}.928 & \cellcolor{gray!15}.936 & \cellcolor{gray!15}.420 & \cellcolor{gray!15}.504 & \cellcolor{gray!15}.612 & \cellcolor{gray!15}.660 & \cellcolor{gray!15}.524 & \cellcolor{gray!15}.424 & \cellcolor{gray!15}\textbf{.658} {\color{red}↑88.5\%} \\
        & \cellcolor{gray!15}-- & \cellcolor{gray!15}\checkmark & \cellcolor{gray!15}.984 & \cellcolor{gray!15}.568 & \cellcolor{gray!15}.968 & \cellcolor{gray!15}1.00 & \cellcolor{gray!15}.640 & \cellcolor{gray!15}.516 & \cellcolor{gray!15}.984 & \cellcolor{gray!15}.592 & \cellcolor{gray!15}.712 & \cellcolor{gray!15}.628 & \cellcolor{gray!15}\textbf{.760} {\color{red}↑117.8\%} \\
        \bottomrule
    \end{tabular}
    }
    \label{tab:further_analysis}
\end{table*}

\vspace{+0.2cm}
\noindent \textbf{Question 1: How much the ImageGen-CoT (via prompt) improves model's performance.}

To verify the effectiveness of \ModelName (via prompt), we compare the model’s performance with and without generating \ModelName before ImageGen via prompt on CoBSAT~\cite{zeng2024can} and Dreambench++~\cite{peng2024dreambench++}. Since CoBSAT includes 10 tasks, we calculate the average score to represent overall performance. Similarly, for Dreambench++, we compute the average score across its tasks.
\vspace{+0.1cm}

\noindent \textbf{Results} As shown in Tables~\ref{tab:main_result_on_cobsat} and \ref{tab:main_results_on_dreambench++}, integrating \ModelName through prompting yields consistent improvements across benchmarks. On CoBSAT, SEED-LLaMA's average score improves from 0.254 to 0.283 (+11.4\% relative gain), while SEED-X shows a more substantial improvement from 0.349 to 0.439 (+25.8\%). The trend persists on Dreambench++, where SEED-X achieves a 84.6\% relative improvement(0.188 → 0.347) compared to its baseline.  These results highlight the effectiveness of incorporating \ModelName in enhancing model performance. However, the SEED performance on Dreambench remains unchanged. This is attributed to its limited comprehension capabilities, which result in unreasonable and disorganized ImageGen-CoT outputs. To address this, we fine-tune the model using our collected ImageGen-CoT datasets, enabling higher-quality generation. More details are provided below.

\vspace{+0.2cm}
\noindent \textbf{Question 2: To what extent does the performance of the model improve after fine-tuning with the \ModelName dataset?}

To further enhance model performance, we fine-tuned both SEED-LLaMA~\cite{ge2023making} and SEED-X~\cite{ge2024seed} on the \ModelName dataset, which was collected using an automatic dataset construction pipeline. The \ModelName dataset consists of two components: the first part focuses on teaching the model how to generate \ModelName text, while the second part trains the model to generate images based on the generated \ModelName text. As described in sec.\ref{sec:training}, our primary goal is to improve the model's capabilities in generating high-quality \ModelName. To this end, we fine-tune the models using Part I of the \ModelName Dataset by default. We compare the performance of our fine-tuned model with its version using ImageGen-CoT (via prompt) and the standard version.

\vspace{+0.1cm}
\noindent \textbf{Results} As shown in Table~\ref{tab:main_result_on_cobsat}, SEED-LLaMA and SEED-X fine-tuned with \ModelName Dataset achieve improvements of +2.8\% (0.283 → 0.291) and +49.9\% (0.439 → 0.658), compared to generating ImageGen-CoT via prompting, respectively. What's more, they even outperform themselves fine-tuned with GT Images by +2.8\% (0.283 → 0.291) and +11.1\% (0.592 → 0.658). Additionally, on the Dreambench++ benchmark, SEED-LLaMA fine-tuned with \ModelName Dataset shows an improvement of +29.5\% (0.078 → 0.101) in CP·PF score, while SEED-X achieves a +16.1\% gain (0.347 → 0.403). These strong results on COBSAT and Dreambench++ underscore the effectiveness and generalizability of our collected \ModelName dataset in enhancing model reasoning and understanding abilities.

\vspace{+0.2cm}
\noindent \textbf{Question 3: Can we invest more time in inference time to improve the performance?}

To further enhance model performance, we explore various test-time scaling strategies. we implement a Best-of-N approach where the model generates multiple image variations, with ground-truth metric evaluation (pass@N). As a baseline approach, we first experiment with the vanilla SEED-X model, generating multiple images by varying the seed values. We then investigate three advanced scaling strategies using SEED-X fine-tuned with ImageGen-CoT dataset: (1) Multi-Chain Scaling, which generates multiple distinct ImageGen-CoT chains, with each chain producing an image; (2) Single-Chain Scaling, which produces multiple image variations from a single ImageGen-CoT chain; and (3) Hybrid Scaling, a novel approach that combines the strengths of both strategies by first generating multiple ImageGen-CoT chains and then producing multiple image variations for each chain. For each paradigm, we systematically evaluate scalability by generating 2, 4, 8, and 16 outputs. For Hybrid Scaling, we implement specific configurations: Hybrid@16 uses 4 ImageGen-CoT chains with 4 images per chain; Hybrid@8 explores two alternatives (2 chains × 4 images or 4 chains × 2 images); Hybrid@4 employs 2 chains × 2 images; and Hybrid@2 tests either 2 chains × 1 image or 1 chain × 2 images. Due to the significant scale difference, we visualize the latter strategy here.

\vspace{+0.1cm}
\noindent \textbf{Results}

As shown in Figure~\ref{fig:test-time-scale-up}, our experiments reveal three key insights. First, the Vanilla SEED-X@16 baseline (0.67 on CobSAT, 0.312 on Dreambench++) underperforms even the simplest scaling strategies (e.g., 0.747 on CobSAT@2), highlighting the necessity of ImageGen-CoT integration. Second, Multi-Chain Scaling matches Single-Chain Scaling in performance, proving that generating diverse reasoning paths is as effective as varying outputs from a single chain. Finally, Hybrid Scaling consistently achieves the highest scores across benchmarks. At N=16, Hybrid Scaling improves CobSAT performance to 0.909 (1.9\% over Single-Chain) and Dreambench++ to 0.543 (0.8\% higher than Single-Chain). The integration of ImageGen-CoT enables effective bidirectional scaling across both comprehension and generation dimensions. This dual-axis scalability suggests new pathways for optimizing MLLM performance in complex multimodal tasks.

\vspace{+0.2cm}
\noindent \textbf{Qualitative Results}
We further validate the effectiveness of our proposed methods through visualization. Figures~\ref{fig:visible_result_on_cobsat_and_dreambench++} showcase the generation results from SEED-X under different configurations: baseline SEED-X, SEED-X with ImageGen-CoT (via prompting), and SEED-X fine-tuned with the ImageGen-CoT dataset. As shown in the top of Figure~\ref{fig:visible_result_on_cobsat_and_dreambench++}, baseline SEED-X (b) generates a basic book shape but misses the implicit "lace" attribute. With ImageGen-CoT prompting (c), the model's weak comprehension leads to poor ImageGen-CoT quality and even degraded generation quality. After fine-tuning with ImageGen-CoT dataset (d), with help of ImageGen-CoT, the model first successfully infers the shared attribute "lace" in CoT text and then generates the correct image - a book made of lace. Similarly, as shown in the bottom of Figure~\ref{fig:visible_result_on_cobsat_and_dreambench++}, baseline SEED-X (b) only generates a simple egg with an open mouth, ignoring key requirements like "on stone", "in garden", and similar expression (sad with upturned closed mouth). With ImageGen-CoT prompting (c), while the egg is placed on stone, it lacks both the required facial expression and garden environment. After fine-tuning (d), the model successfully understands all task requirements and generates a complete scene with an egg properly placed on a stone in a garden setting, maintaining similar facial features to the input. These qualitative results visually demonstrate the effectiveness of ImageGen-CoT and its corresponding dataset in enhancing model comprehension and generation capability, particularly in handling complex tasks that require attention to detail and scene understanding.
\section{Further Analysis}
\subsection{The principles behind ImageGen-CoT contribute to enhancing the model’s performance.}

As described above, our proposed method, ImageGen-CoT, significantly enhances model performance on T2I-ICL tasks. To better understand why \ModelName improves performance, we hypothesize that ‘a better understanding leads to better generation.’ Specifically, we believe that \ModelName enhances the comprehension capabilities of Unified-MLLMs. To quantitatively assess the model’s comprehension ability, we have the model generate a text description for the next image, as indicated by the ‘Gen\_mode’ label in Table~\ref{tab:further_analysis}. Then we conduct a series of experiments to validate this hypothesis. 

\vspace{+0.1cm}
\noindent \textbf{Results: } The results in Table~\ref{tab:further_analysis} demonstrate that integrating ImageGen-CoT significantly enhances model comprehension capabilities. When applied via prompting, SEED-X’s text generation mode (Txt) achieves substantial gains, with the average score improving from 0.174 to 0.457. Fine-tuning with the ImageGen-CoT dataset further amplifies this advantage, elevating the text mode to a remarkable average score of 0.760 (vs. SEED-X’s baseline of 0.174). Notably, enhanced comprehension also improves image generation (Img): SEED-X with ImageGen-CoT via prompt raises the average score from 0.349 to 0.439, while fine-tuning further boosts it to 0.658. This aligns with our hypothesis: ``a better understanding leads to better generation.''

\vspace{+0.2cm}
\subsection{Main obstacles in T2I-ICL}
In this section we further discuss the main obstacles in T2I-ICL. We identify two primary challenges: First, as shown in Table~\ref{tab:further_analysis}, SEED-X's text generation mode (Txt) demonstrates relatively low performance scores (0.174), highlighting its difficulties in comprehending complex T2I-ICL instructions. Second, the image generation capabilities remain a bottleneck - notably, SEED-X fine-tuned with ImageGen-CoT dataset shows lower performance in image generation mode compared to text mode, indicating that while understanding may improve, translating this understanding into accurate image generation remains challenging.

\section{Conclusion}
In this work, we propose a novel framework that enhances
Unified MLLMs’ performance on T2I-ICL tasks by incorporating CoT reasoning before ImageGen. To further improve their performance, we develop an automatic pipeline
to curate high-quality ImageGen-CoT datasets and fine-tune these models. Our extensive experiments demonstrate that our method significantly improves model performance, with SEED-X achieving up to 80\% gains on T2I-ICL tasks after fine-tuning. We further explore test-time scaling strategies and propose a hybrid approach that combines multiple reasoning chains with diverse image generation. Our work establishes a novel paradigm for enhancing MLLMs’ capabilities in handling complex multimodal generation tasks.
\clearpage
\maketitlesupplementary
\setcounter{section}{0}
\setcounter{table}{0}
\setcounter{figure}{0}

\renewcommand{\thesection}{S-\Alph{section}}   
\renewcommand {\thetable} {S-\arabic{table}}
\renewcommand {\thefigure} {S-\arabic{figure}}
\pagebreak

\section*{Overview}
\noindent In this supplementary material, we present more details and more experimental results that are not included in the main paper. The contents include:

\begin{itemize}
    \item A detailed introduction to CoBSAT~\cite{zeng2024can} and Dreambench++\cite{peng2024dreambench++} in Sec.~\ref{sec:dataset_inro}.\\
    
    \item Additional details on the experimental setup in Sec.~\ref{sec:experimental setups}. \\

    \item More experimental results in Sec.~\ref{sec:experimental results}. \\

    \item Effectiveness of Iterative Refinement Strategy in Data Construction in Sec.~\ref{Sec:Eval on pipeline}. \\

    \item Automatic Dataset Construction Pipeline On CoBSAT ~\ref{Sec:Pipeline On CoBSAT}.

\end{itemize}

\section{Dataset Details}
\label{sec:dataset_inro}
\noindent \textbf{CoBSAT}: CoBSAT~\cite{zeng2024can} is a comprehensive benchmark dataset designed specifically to evaluate Text-to-Image In-Context Learning (T2I-ICL) capabilities of Multimodal Large Language Models (MLLMs). The dataset consists of ten distinct tasks across five thematic areas, with each task carefully structured to assess different aspects of T2I-ICL performance. The benchmark is organized into two main categories: object-inference tasks and attribute-inference tasks. In object-inference tasks, models must infer the correct object from demonstrations while being given explicit attributes in the text prompt. Conversely, in attribute-inference tasks, models are provided with the object in the text prompt and must infer the appropriate attribute from the demonstrations. This dual structure enables a thorough evaluation of MLLMs' ability to learn and generalize from multimodal in-context examples.

\noindent \textbf{Dreambench++}: 
Dreambench++~\cite{peng2024dreambench++} is a comprehensive benchmark for evaluating personalized text-to-image generation models. It features three key advantages: 1) Human-aligned evaluation through carefully designed GPT prompting that achieves over 79\% agreement with human assessments; 2) Fully automated evaluation process that eliminates the need for time-consuming manual evaluation; and 3) A diverse dataset containing 150 images and 1,350 prompts across various categories including animals, humans, objects and styles. The benchmark evaluates two fundamental aspects of personalized image generation: concept preservation and prompt following capabilities. 

\section{Detailed Experimental Setup}
\label{sec:experimental setups}

\subsection{CoBSAT}
\noindent\textbf{Data Split.} Following CoBSAT's default settings, we split the predefined lists of text inputs ($X$) and latent variables ($\Theta$) into training and testing subsets with a 1:1 ratio, ensuring the test set contains completely unseen prompts and attributes. For training, we generate 300 samples per task by enumerating all possible combinations of $\theta \in \Theta_{\text{train}}$ and $(x_n)_{n=1}^{N+1} \in X_{\text{train}}^{N+1}$, resulting in 3,000 training samples across 10 tasks. For evaluation, we randomly sample 250 prompts per task from $\theta \in \Theta_{\text{test}}$ and $(x_n)_{n=1}^{N+1} \in X_{\text{test}}^{N+1}$, yielding a total of 2,500 test samples.

\noindent\textbf{Training Strategy.} For model training, we fine-tune both SEED-LLaMA and SEED-X using LoRA. Specifically, SEED-LLaMA is fine-tuned with rank=64, $\alpha$=16, learning rate=1e-4 for 1 epoch, while SEED-X uses rank=64, $\alpha$=64, learning rate=1e-4 for 1 epoch.

\subsection{Dreambench++}
\noindent\textbf{Data Split.} To prevent subject overlap in evaluation, we split the dataset by subjects, with 60\% subjects (90 subjects, resulting in 810 samples) for training and 40\% subjects (60 subjects, resulting in 540 samples) for testing.

\noindent\textbf{Training Strategy.} For Dreambench++, SEED-LLaMA is fine-tuned using LoRA with rank=64, $\alpha$=16, learning rate=1e-4 for 5 epochs, while SEED-X uses rank=64, $\alpha$=64, learning rate=1e-4 for 1 epoch.

\begin{table*}[t]
    \centering
    \caption{\textbf{Main results on CoBSAT benchmark.} "FT w/ GT Image" denotes fine-tuning with ground truth images, while "FT w/ ImageGen-CoT" represents fine-tuning with our ImageGen-CoT dataset. The results demonstrate that ImageGen-CoT significantly improves model performance, with relative improvements over baseline model shown in red.}
    \vspace{+10pt}
    \resizebox{\linewidth}{!}{
    \begin{tabular}{l|c c c c c|c c c c c|l}
        \toprule
        \multirow{2}{*}{\textbf{Method}} & \multicolumn{5}{c|}{\textbf{Object-Inference Task}} & \multicolumn{5}{c|}{\textbf{Attribute-Inference Task}} & \multirow{2}{*}{\textbf{Avg.↑}} \\
        \cmidrule(lr){2-6} \cmidrule(lr){7-11}
        & Color-I & Bkg-I & Style-I & Action-I & Texture-I & Color-II & Bkg-II & Style-II & Action-II & Texture-II & \\
        \midrule
        SEED-LLaMA & .616 & .216 & .272 & .592 & .112 & .088 & .168 & .192 & .220 & .056 & .254 \\
        + \ModelName (via Prompt) & .700 & .276 & .300 & .408 & .084 & .176 & .292 & .272 & .192 & .132 & .283 \\
        + FT w/ GT Image & .632 & .272 & .352 & .540 & .128 & .164 & .200 & .256 & .172 & .112 & .283 \\
        + FT w/ \ModelName Dataset (Part1) & .620 & .368 & .384 & .424 & .060 & .192 & .288 & .208 & .216 & .148 & .291 \\
        \rowcolor{gray!15} + FT w/ \ModelName Dataset (All Part) & .716 & .432 & .436 & .420 & .200 & .168 & .380 & .256 & .216 & .248 & \textbf{.347} {\color{red}↑36.6\%} \\
        \midrule
        SEED-X & .796 & .412 & .316 & .596 & .240 & .176 & .344 & .260 & .252 & .104 & .349 \\
        + \ModelName (via Prompt) & .724 & .440 & .660 & .784 & .216 & .312 & .472 & .228 & .320 & .240 & .439 \\
        + FT w/ GT Image & .936 & .712 & .896 & .860 & .468 & .280 & .324 & .388 & .636 & .424 & .592 \\
       + FT w/ \ModelName Dataset (Part1) & .884 & .692 & .928 & .936 & .420 & .504 & .612 & .660 & .524 & .424 & \textbf{.658} \\
         \rowcolor{gray!15} + FT w/ \ModelName Dataset (All Part) & .832 & .596 & .840 & .892 & .484 & .384 & .548 & .572 & .608 & .500 & .626 {\color{red}↑79.4\%} \\
        \bottomrule
    \end{tabular}
    }
    \label{tab:main_result_on_cobsat}
\end{table*}

\begin{table*}[t]
\caption{\textbf{Evaluation results on Dreambench++ benchmark.} CP refers to concept preservation and PF refers to prompt following metrics. "FT" stands for fine-tuning. The relative gains over baseline model are shown in red.}
\vspace{2pt}
\resizebox{\linewidth}{!}{
\begin{tabular}{l|c c c c c|c c c c|l}
\toprule
\multirow{2}{*}{\textbf{Method}} & \multicolumn{5}{c|}{\textbf{Concept Preservation}} & \multicolumn{4}{c|}{\textbf{Prompt Following}} & \multirow{2}{*}{\textbf{CP$\cdot$PF↑}} \\
\cmidrule(lr){2-6}
\cmidrule(lr){7-10}
& \texttt{Animal} & \texttt{Human} & \texttt{Object} & \texttt{Style} & Overall & \texttt{Photorealistic} & \texttt{Style} & \texttt{Imaginative} & Overall & \\
\midrule
SEED-LLaMA & .436 & .315 & .288 & .381 & .358 & .306 & .202 & .154 & .218 & .078 \\
+ \ModelName (via Prompt) & .390 & .241 & .262 & .346 & .317 & .291 & .211 & .170 & .222 & .078 \\
+ FT w/ \ModelName Dataset (Part1) & .399 & .290 & .271 & .318 & .325 & .348 & .355 & .210 & .310 & .101 \\
\rowcolor{gray!15} + FT w/ \ModelName Dataset (All Part) & .414 & .269 & .243 & .328 & .319 & .408 & .317 & .199 & .334 & \textbf{.107} {\color{red}↑37.2\%} \\
\midrule
SEED-X & .647 & .420 & .526 & .571 & .559 & .346 & .342 & .303 & .337 & .188 \\
+ \ModelName (via Prompt) & .547 & .293 & .369 & .424 & .427 & .862 & .775 & .737 & .817 & .347 \\
+ FT w/ \ModelName Dataset (Part1) & .549 & .410 & .403 & .432 & .458 & .922 & .851 & .846 & .881 & \textbf{.403}\\
\rowcolor{gray!15} + FT w/ \ModelName Dataset (All Part) & .511 & .358 & .424 & .303 & .421 & .926 & .910 & .870 & .906 & .384 {\color{red}↑104.2\%} \\
\bottomrule
\end{tabular}
}
\label{tab:main_results_on_dreambench++}
\end{table*}

\section{More experimental results}
\label{sec:experimental results}
As described in the main paper, the ImageGen-CoT dataset comprises two distinct components. The first component focuses on training the model to generate ImageGen-CoT text, while the second component teaches the model to generate images based on the generated ImageGen-CoT text. While our main paper primarily focused on training using Part I of the dataset, here we extend our experiments by utilizing the complete dataset for comprehensive evaluation. As presented in Tables ~\ref{tab:main_result_on_cobsat} and ~\ref{tab:main_results_on_dreambench++}, we conducted comprehensive experiments using both parts of the ImageGen-CoT dataset. On the CoBSAT benchmark, SEED-LLaMA fine-tuned with the complete ImageGen-CoT dataset achieved a significant performance gain of +36.6\% (0.254 → 0.347) compared to the baseline model. Similarly, SEED-X demonstrated remarkable improvement with a +79.4\% increase (0.349 → 0.626) over its baseline performance. For the Dreambench++ benchmark, training with the complete dataset resulted in even more pronounced improvements. SEED-LLaMA showed a +37.2\% gain (0.078 → 0.107) in CP·PF score, while SEED-X achieved a substantial +104.2\% improvement (0.188 → 0.384). These comprehensive results demonstrate that utilizing the complete ImageGen-CoT dataset can still significantly improve model performance.

\section{Effectiveness of Iterative Refinement Strategy in Data Construction}
\label{Sec:Eval on pipeline}

We evaluate the effectiveness of our iterative refinement strategy on both CoBSAT and Dreambench++ datasets. As demonstrated in Table \ref{tab:effectiveness}, the proposed strategy yields consistent improvements across all evaluation metrics. Specifically, on the CoBSAT dataset, our method achieves improvements of 1.1\%, 2.9\%, and 2.0\% in object inference, attribute inference, and overall score, respectively. For Dreambench++, the refinement strategy enhances prompt following (PF) by 0.9\% and concept preservation (CP) by 4.7\%, resulting in a substantial 4.7\% improvement in the combined PF*CP metric. These comprehensive results validate that our iterative refinement approach significantly enhances the quality of the constructed dataset.

\begin{table}[t]
\caption{Performance comparison of data construction with and without iterative refinement.}
\label{tab:effectiveness}
\begin{subtable}[t]{0.48\textwidth}
    \centering
    \caption{Results on CoBSAT}
    \begin{tabular}{l|ccc}
    \hline
    Method & Object & Attribute & Overall \\
    \hline
    w/o Iterative Refine & 0.782 & 0.704 & 0.743 \\
    w/ Iterative Refine & 0.793 & 0.733 & \textbf{0.763} \\
    \hline
    \end{tabular}
\end{subtable}
\hfill
\begin{subtable}[t]{0.48\textwidth}
    \centering
    \caption{Results on Dreambench++}
    \begin{tabular}{l|ccc}
    \hline
    Method & PF & CP & PF*CP \\
    \hline
    w/o Iterative Refine & 0.937 & 0.470 & 0.442 \\
    w/ Iterative Refine & 0.946 & 0.517 & \textbf{0.489} \\
    \hline
    \end{tabular}
\end{subtable}
\end{table}

\section{Automatic Dataset Construction Pipeline On CoBSAT}
\label{Sec:Pipeline On CoBSAT}
On CoBSAT, we initially adopted the same method as DreamBench++. However, we found that the self-boosting approach underperformed due to the inherent complexity of CoBSAT, which requires the model to infer implicit visual-semantic relationships—posing a significant challenge to the model’s reasoning capability. To solve this challenge, we sampled multiple text prompts from the MLLM and selected the best prompts using the self-consistency method. However, this method cannot be directly applied to CoBSAT. Self-consistency is commonly used in mathematical problem solving, where text answers are precise (e.g., numbers or options) and consistency can be directly evaluated using string matching. In contrast, CoBSAT involves long and complex text prompts, making direct string-based consistency evaluation infeasible. 

The pipeline proceeds as follows: We first sample multiple chains of thought (CoT) from the MLLM. These CoTs are then used, along with the input sequence context, to generate multiple text prompts. Formally, let the CoT sampled from the MLLM be denoted as \( \text{cot}_t^i \), where \( i = 0, 1, \dots, M-1 \), and \( M \) is the number of sampled CoTs. Each CoT, combined with the input sequence \( x \), is used to construct a corresponding text prompt \( p_t^i \) as:

\begin{equation}
    p_t^i = \mathcal{F}(\text{cot}_t^i, x), \quad i = 0, 1, \dots, M-1,
    \label{eq:cot_to_prompt}
\end{equation}

where \( \mathcal{F} \) represents generating text prompts based on the CoT and input sequence context.

Next, we convert each text prompt into a vector representation using a text embedding model \( \mathcal{E} \):

\begin{equation}
    v_t^i = \mathcal{E}(p_t^i), \quad i = 0, 1, \dots, M-1,
    \label{eq:text_embedding}
\end{equation}

where \( v_t^i \in \mathbb{R}^d \) is the embedding vector of the \( i \)-th prompt.

The similarity \( S_{ij} \) between two prompts is then measured using the inner product of their vector representations:

\begin{equation}
    S_{ij} = \langle v_t^i, v_t^j \rangle = v_t^i \cdot v_t^j,
    \label{eq:inner_product}
\end{equation}

where \( \langle \cdot, \cdot \rangle \) denotes the inner product.

The average similarity for each prompt \( p_t^i \) is computed as:

\begin{equation}
    \bar{S}_i = \frac{1}{M-1} \sum_{\substack{j=0 \\ j \neq i}}^{M-1} S_{ij}.
    \label{eq:average_similarity}
\end{equation}

Finally, the prompt \( p_t^{i^*} \) with the highest average similarity is selected as the best candidate:

\begin{equation}
    i^* = \arg\max_i \bar{S}_i.
    \label{eq:best_prompt}
\end{equation}

The selected prompt \( p_t^{i^*} \) is then used to generate the image, which is considered the best image. Simultaneously, its corresponding CoT is also identified as the best CoT. The CoT text and the generated image are then concatenated to form the \ModelName dataset.
{
    \small
    \bibliographystyle{ieeenat_fullname}
    \bibliography{main}
}

\end{document}